\documentclass[runningheads]{llncs}
\usepackage[T1]{fontenc}
\usepackage{graphicx}
\usepackage{adjustbox}
\usepackage{booktabs}
\usepackage{comment}
\usepackage{siunitx}
\usepackage{float}
\usepackage{hyperref}
\usepackage{wrapfig}
\usepackage{ragged2e}

\usepackage{multirow}

\usepackage{cleveref}
\crefformat{section}{\S#2#1#3} % see manual of cleveref, section 8.2.1
\crefformat{subsection}{\S#2#1#3}
\crefformat{subsubsection}{\S#2#1#3}

\begin{document}
\title{Towards Efficient and Explainable Hate Speech Detection via Model Distillation}
\titlerunning{Efficient and Explainable Hate Speech Detection via Model Distillation}

\author{Paloma Piot\orcidID{0000-0002-7069-3389} \and
Javier Parapar\orcidID{0000-0002-5997-8252}}

\institute{IRLab, CITIC, Universidade da Coruña, Spain\\ \email{paloma.piot@udc.es} \email{javier.parapar@udc.es}} 

\maketitle              % typeset the header of the contribution
\begin{abstract}

Automatic detection of hate and abusive language is essential to combat its online spread. Moreover, recognising and explaining hate speech serves to educate people about its negative effects. However, most current detection models operate as black boxes, lacking interpretability and explainability. In this context, Large Language Models (LLMs) have proven effective for hate speech detection and to promote interpretability. Nevertheless, they are computationally costly to run. In this work, we propose distilling big language models by using Chain-of-Thought to extract explanations that support the hate speech classification task. Having small language models for these tasks will contribute to their use in operational settings. In this paper, we demonstrate that distilled models deliver explanations of the same quality as larger models while surpassing them in classification performance. This dual capability —classifying and explaining— advances hate speech detection making it more affordable, understandable and actionable.\footnote{Our code, models and prompts are available at \url{https://github.com/palomapiot/distil-metahate}.}

\vspace{3mm}

\textcolor{red}{ This article contains illustrative instances of hateful language.}
\keywords{hate speech  \and explainable AI \and knowledge distillation}
\end{abstract}
\section{Introduction}
\label{sec:intro}

Social media and internet platforms have considerably enhanced global networking and communication. But this rise in connectivity has also exacerbated the propagation of hate speech, which is affecting people globally \cite{pewresearch2021onlineharassment,hickey2023auditing}. The literature consensus \cite{davidsonhatespeech,fountahatespeech,mathewb2020hatexplain,hatelingo2018elsherief,ElSherief2018,silva2016analyzing,hatemm2023} on the hate speech definition frames it as ``\textit{language characterized by offensive, derogatory, humiliating, or insulting discourse \cite{fountahatespeech} that promotes violence, discrimination, or hostility towards individuals or groups \cite{davidsonhatespeech} based on attributes such as race, religion, ethnicity, or gender \cite{hatelingo2018elsherief,ElSherief2018,hatemm2023}}''. In the rest of this paper, we will assume that definition, which highly aligns with the United Nations one \cite{unhatespeech}, differentiating hate from offensive and neutral speech. Regarding the gravity of this phenomenon in the online world, studies reveal that about 30\% of youth experience cyberbullying \cite{kansok2023systematic}, and 46\% of Black and African American adults encounter online racial harassment \cite{adl2024}. 

It was already in 1997 when researchers suggested to use Machine Learning (ML) models to identify different types of harmful messages \cite{10.5555/1867406.1867616}. Since then, there have been efforts for detecting hate speech spanning from linear models like Support Vector Machines (SVM) or Logistic Regressors \cite{waseem-hovy-2016-hateful,davidsonhatespeech,Salminen2018} to more recent state-of-the-art methods based on transformers \cite{Samory2021,10.1145/3583780.3615260,Piot_Martin-Rodilla_Parapar_2024} or LLMs \cite{kumarage2024harnessingartificialintelligencecombat,10.1145/3605098.3635964,plaza-del-arco-etal-2023-respectful}. 

\begin{wrapfigure}{r}{0.34\textwidth} %this figure will be at the right
    \centering
    \includegraphics[width=0.34\textwidth]{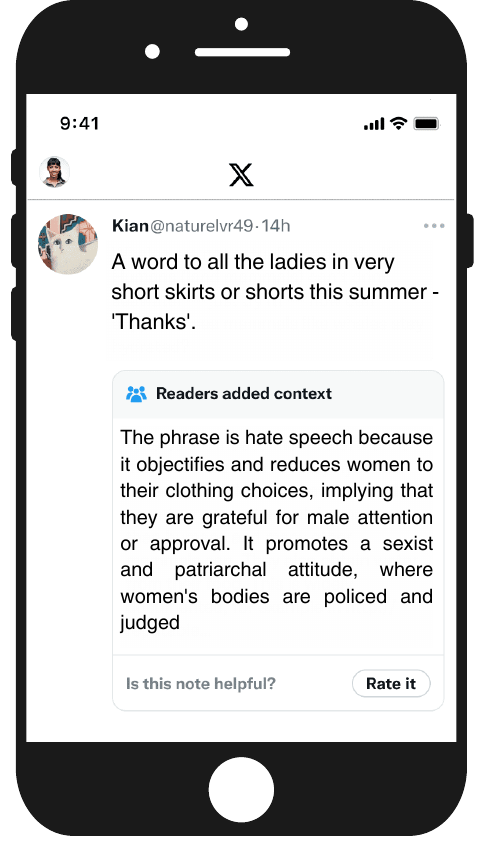}
\caption{X's Community Notes feature for adding post context.}
\label{fig:community-notes}
\end{wrapfigure}

Many ML models have limitations because they operate as black boxes, making them hard to interpret. Also, these models only classify without explaining the reasons. Explaining why content is blocked or flagged as hate speech is essential for transparency and trust in online platforms \cite{Gillespie2018,liu-etal-2019-towards-explainable}. 

The Digital Services Act (DSA) \cite{dsa}, which came into force for all platforms in 2024, is a landmark regulation that sets clear obligations for online platforms, including social media, on content moderation practices. Platforms must provide clear explanations for content removal or restrictions, along with an appeals process for users. Existing efforts for content moderation span from automatic moderation to manual revision of publications, including new approaches like Community Notes by X (formerly Twitter). This feature provides additional information and context about a post. Ideally, automatic detection and explanation of hate speech could be implemented in a similar way (see Figure \ref{fig:community-notes}). Clear natural language rationales help users understand the reasons behind hateful actions, promoting fairness and reducing misunderstandings \cite{10.1145/3616865}. To generate such rationales, LLMs were proposed in Zero-Shot, Few-Shot, or fine-tuned settings \cite{plaza-del-arco-etal-2023-respectful,GarcaDaz2023,Pan2024}. However, although, LLMs excel in reasoning and offer remarkable capabilities, they come with notable challenges, particularly in terms of cost, deployment, and resource requirements \cite{10.1145/3381831,10.1145/3442188.3445922}.

We propose a method that leverages LLM-generated rationales with Few-Shot and Chain-of-Thought (CoT) \cite{10.5555/3600270.3602070} prompts to distil smaller models. This approach retains the advanced capabilities of larger models while being more cost-effective \cite{yang-etal-2023-hare,nirmal2024interpretablehatespeechdetection,Zhang2024}. Our method involves extracting rationales with 70B LLMs using Few-Shot CoT prompting, followed by employing them to fine-tune a smaller model within a multi-task learning framework. This framework enables models to learn both classification (generating labels) and explanation (providing supporting rationales). We evaluate our method against state-of-the-art models and conduct qualitative analysis with human reviewers to assess the quality of the rationales.

Our contributions are the following: (i) we propose a hate speech detection approach that uses sample data from 36 diverse human-curated datasets, enhancing its generalisation capabilities; (ii) we fine-tune a small language model, transferring knowledge from a larger one, to predict both labels and rationales using a multi-task learning framework; (iii) we evaluate the classification task against four state-of-the-art models and conduct a human evaluation to assess the quality of the generated rationales.

\section{Related Work}

Most traditional approaches for hate speech classification have focused on predicting whether a piece of text is hate speech or not \cite{de-gibert-etal-2018-hate,Samory2021,10.1145/3583780.3615260}. Additionally, some works have addressed multicategory and multilabel approaches \cite{waseem-hovy-2016-hateful,davidsonhatespeech}. Recent studies have also explored the use of LLMs for hate speech classification \cite{wang2022toxicitydetectiongenerativepromptbased,kumarage2024harnessingartificialintelligencecombat}. These methods require large amounts of data to achieve generalisation and often lack explainability, making it difficult to understand why a text is labelled as hate speech.

More recently, some studies have attempted to add explanations to hate speech posts, typically by marking the spans of text where the hate speech is present \cite{mathewb2020hatexplain} or providing vague explanations that lack sufficient detail \cite{sap-etal-2020-social,caselli-etal-2020-feel}. Other researchers have explored generating explanations using LLMs \cite{yang-etal-2023-hare,Zhang2024,nirmal2024interpretablehatespeechdetection}. For example, Nirmal et al. \cite{nirmal2024interpretablehatespeechdetection} propose using LLMs to extract features from input texts to train BERT classifiers, aiming to improve interpretability. Their framework maintains classification performance while enhancing the model's explainability. While these studies have improved over traditional binary hate speech classification approaches, they often use explanations to enhance the detection process without assessing the quality of the explanations.

These new approaches using big LLM are very costly to run. This opens up an avenue for knowledge distillation, a mechanism of transferring expertise from a large model to a smaller, more affordable one. This technique started with BERT and transformer-based models \cite{sanh2020distilbertdistilledversionbert,he-etal-2021-distiller}. However, with the rise of LLMs, this method has become particularly important, as it enables LLMs' capabilities to be transferred to smaller models, improving efficiency, speed, and scalability while reducing costs and resource requirements. Hsieh et al. \cite{hsieh-etal-2023-distilling} propose a novel approach to generate step-by-step explanations with LLMs and distil this knowledge into smaller models. Other recent works have also focused on similar methods to improve LLMs' reasoning skills \cite{wang2023pintofaithfullanguagereasoning} and enhance their performance in commonsense reasoning and arithmetic tasks \cite{magister-etal-2023-teaching,ho-etal-2023-large}. 

In this line of distilling LLMs, Yang et al. propose a method for generating hate speech explanations using LLMs, which are then used to fine-tune other models (knowledge distillation) \cite{yang-etal-2023-hare}. They generate explanations in a zero-shot scenario, and also using human-authored reasoning. The study assesses the classification improvement gained by using these rationales but does not evaluate the quality of the generated explanations. Zhang et al. have distilled LLMs for toxic content detection using Decision-Tree-of-Thought (DToT) prompting \cite{Zhang2024}. Their approach improves the classification accuracy of baseline models, but the quality of the generated rationale is only evaluated by checking if the model generates more than a binary response (toxic yes/no) without assessing the explanation's quality. Our work aligns with this research, focusing on detecting hate speech in social media posts and generating supporting evidence to improve detection. However, we also evaluate the quality of the generated rationales, assessing both their completeness and correctness.

\section{Proposal}

\begin{figure}[h!]
\centering
\includegraphics[scale=0.48]{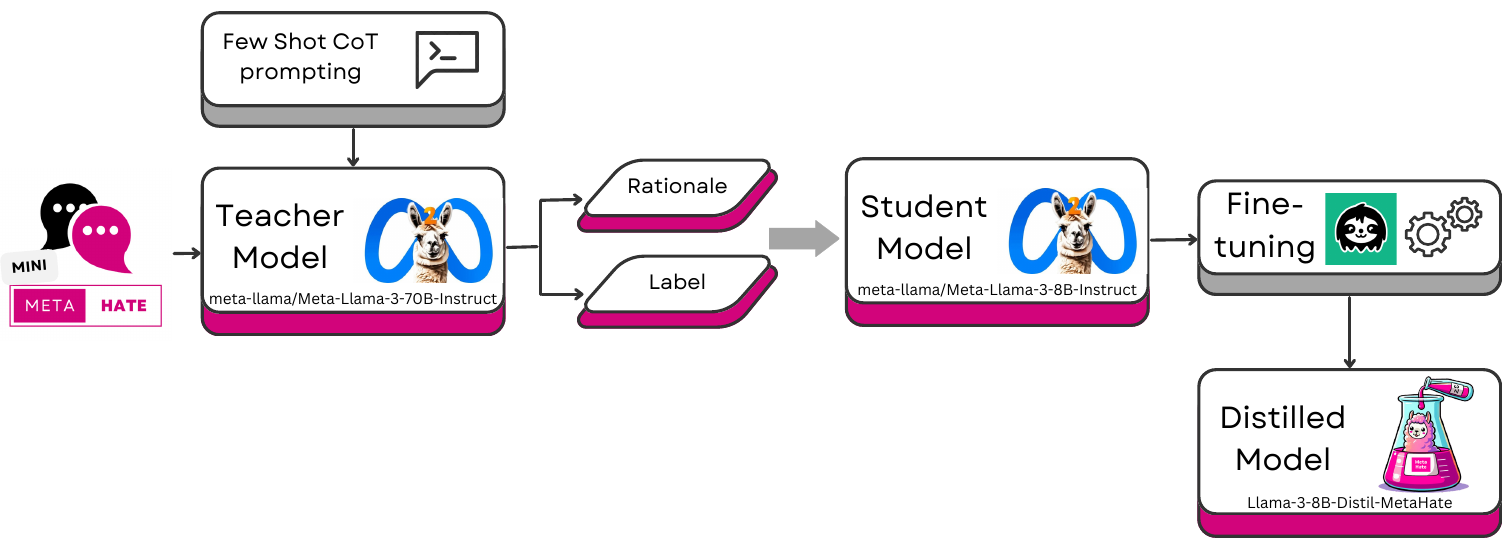}
\caption{LLM Knowledge distillation with CoT overview.}
\label{fig:workflow}
\end{figure}

We propose performing knowledge distillation using the \texttt{Llama-3-70B-Instruct} model as the \textbf{big}, \textbf{teacher} model, transferring its knowledge to the \textbf{small}, \textbf{student} \texttt{Llama-3-8B-Instruct} model \cite{llama3modelcard}. We prepared two hate speech datasets as explained in \cref{sec:datasets}. With the first, we extracted from the teacher model binary labels for hate speech together with the explanations (rationales), using Few-Shot CoT prompting. Then, we transferred the knowledge to the student model and distilled it following the procedure described in \cref{sec:kd}, resulting in \texttt{Llama-3-8B- Distil-MetaHate}. For evaluation, we employed the second dataset, and we tested the \textbf{teacher} model, its smaller version (\textbf{base}), and the new distilled version (\textbf{distilled}). Figure \ref{fig:workflow} shows the experimental setup of our current work.

\subsection{Knowledge Distillation}
\label{sec:kd}

Knowledge Distillation is a machine learning technique designed to transfer knowledge from a large pre-trained model, called \texttt{teacher}, to a smaller \texttt{student} model. The goal of knowledge distillation is to train a smaller, more efficient model to replicate the behaviour of a larger, more complex model. Next, we explain how we performed this process.

\subsubsection{Extract Rationales}
\label{sec:exp_extract_rationales}

The first phase is to extract knowledge from the large model. For this, we employed Chain-of-Thought In-Context Learning \cite{NEURIPS2020_1457c0d6,10.5555/3600270.3602070} to extract rationales from the teacher model. To achieve this, we created a prompt template ${\displaystyle p}$ that explains how to solve the task, focusing on two main points: (1) providing the model with the definition of hate speech, and (2) instructing the model on what to do. Each prompt ${\displaystyle p}$ was structured as a tuple ${\displaystyle (H, [x\textsubscript{p}, r\textsubscript{p}, y\textsubscript{p}])}$, where ${H}$ is the hate speech definition, and ${\displaystyle [x\textsubscript{p}, r\textsubscript{p}, y\textsubscript{p}]}$ is a list of examples containing: an input text ${\displaystyle x\textsubscript{p}}$, its corresponding label ${\displaystyle y\textsubscript{p}}$, and a human-authored rationale ${\displaystyle r\textsubscript{p}}$ explaining why ${\displaystyle x\textsubscript{p}}$ is categorized as ${\displaystyle y\textsubscript{p}}$. Let ${\displaystyle D = \{(x\textsubscript{i}, y\textsubscript{i})\}_{i=1}^N}$ represent our dataset, where each ${\displaystyle x\textsubscript{i}}$ denotes an input and ${\displaystyle y\textsubscript{i}}$ is its corresponding label, we appended each dataset instance ${\displaystyle x\textsubscript{i} \in D}$ to the prompt ${\displaystyle p}$ and used it to instruct the model to generate both the rationale ${\displaystyle \hat{r\textsubscript{i}}}$ and the classification ${\displaystyle \hat{y\textsubscript{i}}}$ for each ${\displaystyle x\textsubscript{i}}$. The rationales included in our context prompt are used to guide the model towards our task objective: generate the dataset's ${\displaystyle D}$ rationales ${\displaystyle \hat{r}\textsubscript{i}}$ and labels ${\displaystyle \hat{y}\textsubscript{i}}$. Although the text's label is already in our data, we generated both the label and an explanation, as matching label generations are more likely to have a correct explanation. This allows us to use them to train our smaller model.

\subsubsection{Train Small Model with Rationales}

After obtaining the labels and rationales, we incorporated them into the training process of our distilled model. Now our dataset ${\displaystyle D}$ also includes the generated rationale ${\displaystyle \hat{r}\textsubscript{i}}$. We only included the samples where the teacher model's label matched the gold label, therefore ${\displaystyle {y\textsubscript{i}} = \hat{y\textsubscript{i}}}$. In this work, we are treating this proposal as a multi-task problem where the model ${\displaystyle f(x\textsubscript{i}) \rightarrow (\hat{y\textsubscript{i}}, \hat{r\textsubscript{i}})}$ learns both to predict labels ${\displaystyle \hat{y\textsubscript{i}}}$ and generate rationales ${\displaystyle \hat{r\textsubscript{i}}}$. The overall objective combines these two tasks, leading to the total optimization goal of ${\mathcal{L} = \alpha \mathcal{L}_{ \text{label}} +}$ ${\beta \mathcal{L}_{\text{rationale}}}$ \cite{hsieh-etal-2023-distilling} where ${\mathcal{L}_{\text{label}}}$ focuses on the label prediction ${\mathcal{L}_{\text{label}} = \frac{1}{N} \sum_{i=1}^{N} \ell(y_i, \hat{y}_{i})}$, where ${\displaystyle \ell()}$ is the cross-entropy loss, and ${\mathcal{L}_{\text{rationale}}}$ is:

\begin{equation}
\label{eq:generation}
\mathcal{L}_{\text{rationale}} = \frac{1}{N} \sum_{i=1}^{N} \sum_{t=1}^{T_i} \ell(f(x_i)_t, \hat{r}_{it})
\end{equation}

\Cref{eq:generation} represents the average loss over ${\displaystyle N}$ samples, where for each sample ${\displaystyle i}$ the model generates a sequence of ${\displaystyle T\textsubscript{i}}$ rationale tokens. The loss function ${\displaystyle \ell(f(x_i)_t, \hat{r}_{it})}$ –cross-entropy– measures the discrepancy between the predicted token ${\displaystyle f(x_i)_t}$ and the true token ${\displaystyle \hat{r}_{it}}$ at each step ${\displaystyle t}$. The goal is to minimize this loss across all tokens and samples to improve the rationale generation. The rationale generation helps the model produce intermediate reasoning steps, which can enhance its label prediction accuracy, i.e. by reasoning step-by-step, the model can better analyse the messages, improving its accuracy in the labelling.

\section{Experiments}
\label{sec:experiments}

Now, we describe the datasets, the models, and the experiments conducted.

\subsection{Datasets}
\label{sec:datasets}

We conducted our experiments on two subsamples of MetaHate \cite{Piot_Martin-Rodilla_Parapar_2024}. MetaHate is a meta-collection of 1.2 million entries, labelled in a binary fashion, compiled from 36 different datasets across various social networks. This made it an ideal choice for our study, given the diversity of the data it contains.

For our task, we created two balanced subsamples from this dataset, with around \num{2000} messages each. Each subsample was balanced in two ways: (1) they contained an equal number of entries labelled as hate and non-hate, and (2) the ratio of posts from each original dataset was maintained, reflecting the source distribution in MetaHate. Due to the high computational cost of inferring and fine-tuning LLMs, we chose to work with these samples instead of using the full collection. Moreover, inspired by the results of Hsieh et al. \cite{hsieh-etal-2023-distilling}, which showed that distilling step-by-step explanations outperforms standard fine-tuning with much less data, we decided that using only part of the dataset was adequate.

\begin{itemize}
    \item {\textbf{MiniMetaHate Distil}}: The first subsample was created to extract rationales and labels with the large model, where \num{2993} samples were inferred. Later, together with the generated explanations, it was used to distil the smaller model, but only using the entries where the gold label was equal to the generated label. This ended up in a \num{2296} sample of corrected classified posts. Size: \num{2296} posts.
    \item {\textbf{MiniMetaHate Eval}}: The second subsample was designed to evaluate the models, focusing on both classification performance and the quality of the generated explanations. This set is used to test and validate the model’s effectiveness. Size: \num{2001} posts.
\end{itemize}

\subsection{Chosen LLMs}
\label{sec:models}

For knowledge distillation, we need a large model with strong reasoning capabilities and a smaller counterpart that can learn from it. Based on recent reports \cite{llama3-report-2024}, we decided to include the most capable open-source models available to date in this experiment.

\begin{itemize}
    \item {\textbf{Llama-3-70B-Instruct (Teacher)} \cite{llama3modelcard}}: This instruction-tuned model is optimized for dialogue use cases, offering exceptional performance in conversational tasks. This model surpasses many available open-source chat models on industry-standard benchmarks, providing superior accuracy and responsiveness. It also demonstrates exceptional ability in reasoning tasks.
    \item {\textbf{Llama-3-8B-Instruct (Student)} \cite{llama3modelcard}}: This is the same model as the Teacher but with 8B parameters instead of 70B. It is known for being fast and cost-effective, making it an ideal choice for learning from a larger model and for use in production environments.
\end{itemize}

\subsection{Experimental Settings}
\label{sec:setup}

Our experimental approach consists of two steps: 

First, we used a large language model (\texttt{Llama-3-70B-Instruct}) and a labelled dataset (``MiniMetaHate Distil'') to generate output labels along with explanations, called rationales, that justify the labels. These rationales are natural language explanations that support the model’s predicted labels. For this, we employed Few-Shot Chain-of-Thought prompting, where 12 human-authored examples were provided. We used \texttt{unsloth} 4-bit pre-quantized \texttt{Llama-3-70-Instruct} model, with a max sequence of \num{4096} and max tokens generation of \num{2048}. To facilitate processing, we asked the model to return a JSON object that includes the hate speech classification decision and a list of tuples with input fragments and their explanations. 

Second, we used the output rationales pairs and classification labels to train a smaller model (\texttt{Llama-3-8B-Instruct}). The rationales provide detailed information about why an input text is assigned a specific label and include relevant knowledge that helps clarify why that label was given \cite{hsieh-etal-2023-distilling}. We fine-tuned \texttt{Llama-3-8B-Instruct} using the \num{2296} inferred instances of ``Mini MetaHate Distil''. During this fine-tuning, the model was provided with both labels and rationales to learn the classification and explanation of hate speech texts. We only included samples where the big model's label matched the gold label, where ${\displaystyle {y\textsubscript{i}} = \hat{y\textsubscript{i}}}$. For fine-tuning, we used Quantized Low-Rank Adaptation (QLoRA) for causal language modelling, with an attention dimension of \num{16} and ${\displaystyle \alpha}$ set to \num{32}. The model was trained with 4-bit quantization for \num{1000} steps, using a learning rate of 2.5e\textsuperscript{-5}. 

\begin{figure}[ht!]
\centering
\includegraphics[scale=0.45]{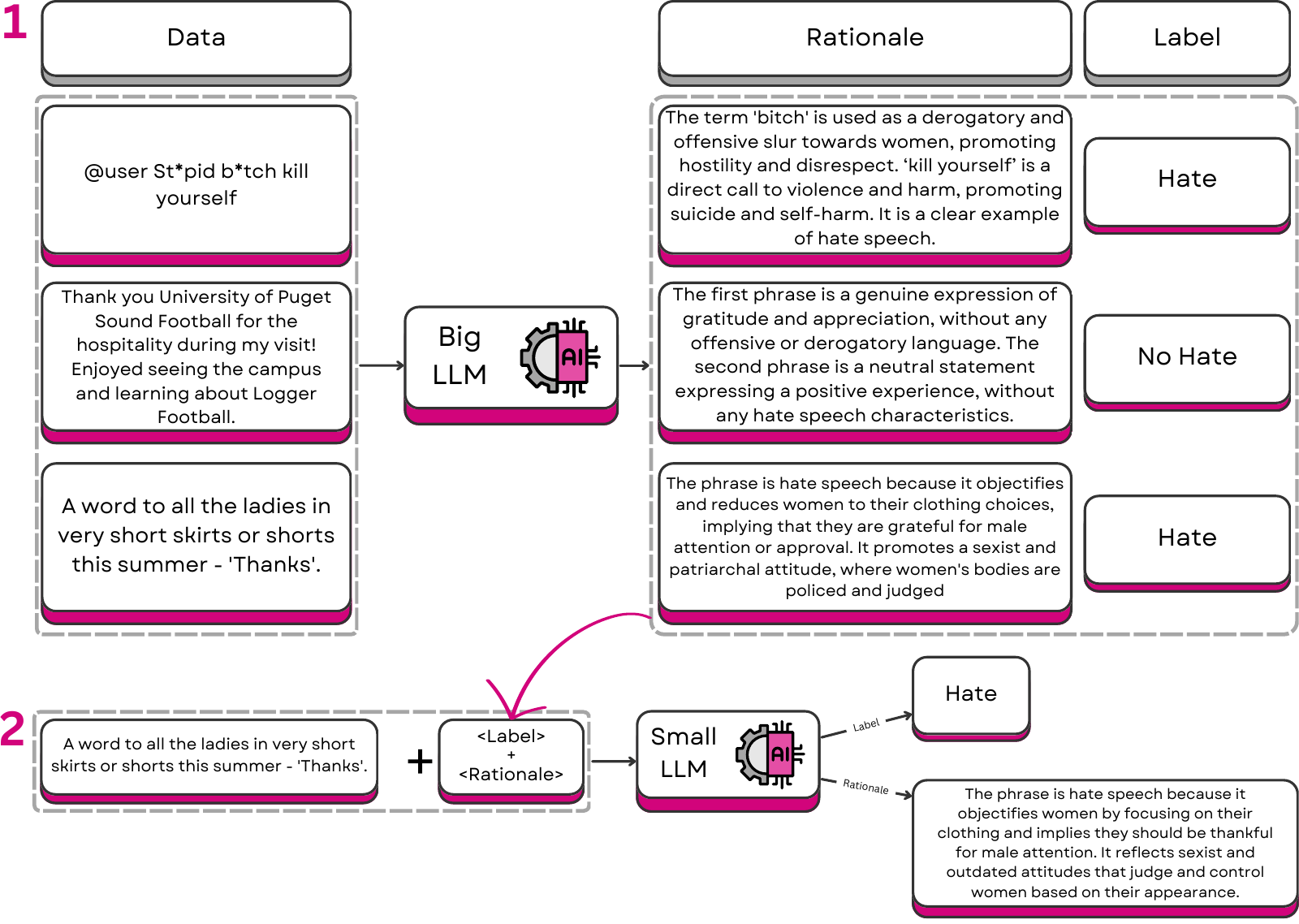}
\caption{Overview of our proposed approach for explaining and detecting hate speech: First, we extract rationales from an LLM using Few-Shot CoT. We then use these rationales, along with the labels, to train a small model within a multi-task learning framework. This enables the small model to explain and detect hate speech effectively.}
\label{fig:overview}
\end{figure}

Figure \ref{fig:overview} illustrates this procedure, showing how the rationales enhance the training of the smaller model by providing context and relevant knowledge that support the labels.

After curating the distilled model, we inferred over ``MiniMetaHate Eval'' using three language models: (1) the big model (\texttt{LLama-3-70B-Instruct}), (2) the small model (\texttt{Llama-3-8B-Instruct}), and (3) the distilled model (\texttt{Llama-3-8B- Distil-MetaHate}). For the big and small models, we provided the same Few-Shot CoT examples used for extracting the rationales in \cref{sec:exp_extract_rationales}. For the distilled model, we only included the task instruction in the context (see Table \ref{tab:instruction}), as it had already been fine-tuned with ``MiniMetaHate Distil'' and had seen over \num{2000} examples during training. We used a max sequence of \num{4096} and max tokens generation of \num{2048}. This step resulted in \num{2001} posts with their explanations and labels for each model.

\begin{table}[ht!]
\centering
\footnotesize
\caption{Task instruction.}
\begin{tabular}{p{12.2cm}}
\toprule
\textbf{Instruction} \\ \midrule
\ttfamily
You must explain why a social media message is hateful or not and then tell me your decision. You must always reply with only a JSON containing one field 'hate\_speech' including a Boolean value (``True'' for hate speech messages, ``False'' for neutral ones); and a field 'explanations' containing a list with each message phrase and its corresponding explanation. Do not include text outside the JSON. 

This is the definition of hate speech: ``language characterised by offensive, derogatory, humiliating, or insulting discourse that promotes violence, discrimination, or hostility towards individuals or groups based on attributes such as race, religion, ethnicity, or gender''. 

Generate step-by-step explanation for: <Message> message </Message> \\
\bottomrule
\end{tabular}
\label{tab:instruction}
\end{table}

To evaluate its quality, we proposed two experiments: the hate speech classification task and the explanation task.

\subsection{Experiment 1: Classification Task}

Our distilled model is trained to classify hate speech and provide explanations. To assess its classification performance, along with the base LLMs, and compare the results to established methods, we selected four state-of-the-art models for hate speech detection as baselines:

\begin{itemize}
    \item \textbf{MetaHateBERT} \cite{Piot_Martin-Rodilla_Parapar_2024}: We chose this model because of its training on a large dataset from multiple social networks, providing diverse data. To ensure no overlap with our MiniMetaHate subsamples, we fine-tuned it again, ensuring the training sample did not include any posts that are in MiniMetaHate subsamples. We followed the same steps as the original publication and fine-tuned \texttt{bert-base-uncased} model for 3 epochs with a learning rate of 5e\textsuperscript{-5} and a batch size of \num{32}.
    \item \textbf{HateBERT} \cite{caselli-etal-2021-hatebert}: This model is a well-established baseline in the hate speech community, trained on Reddit data (RAL-E dataset). Its proven effectiveness in hate speech detection makes it a valuable comparison.
    \item \textbf{HateXplain} \cite{mathewb2020hatexplain}: Trained on the HateXplain dataset, this model detects both hate speech and offensive content, and neutral posts. Despite our binary classification focus, we included it for its strong acceptance in the community. We mapped offensive content to non-hate for our purposes.
    \item \textbf{Perspective API} \cite{10.1145/3534678.3539147}: We selected Perspective API for its widespread use in evaluating online toxicity. The API provides a toxicity score between 0 and 1, therefore, for obtaining a binary classification decision from the score, we use a threshold of 0.5.
\end{itemize}

\subsection{Experiment 2: Explanation Task}
\label{sec:annotation}

As we mentioned, our work involves distilling the model for classifying hate speech and generating rationales to support the classification decision. Therefore, for evaluating the generated rationales, we performed a human evaluation on a \num{100} entries subsample per model. We recruited three annotators with experience in hate speech detection and explainability to annotate a sample of the generated responses. Next, we present the annotation guidelines and procedure performed.

\subsubsection{Hate Speech Definition}

We framed our work under hate speech, as defined in section \cref{sec:intro}, differentiating hate speech from non-hate and offensive speech. We define offensive speech as \textit{posts containing any form of unacceptable language (profanity) or a targeted offence, which can be veiled or direct; this includes insults, threats, and posts containing profane language or swear words \cite{zampieri-etal-2019-predicting}}.

\subsubsection{Annotation process}

First, a group of experts defined the annotation guidelines, identifying two crucial aspects: completeness and correctness.

\begin{itemize}
    \item \textbf{Completeness} refers to whether an annotation fully captures all instances of hate speech within a post. It ensures that every hateful sentence or fragment in the original post is addressed, making sure nothing is overlooked. This aspect is evaluated at the \textbf{post level} using \textit{true} if all hate sentences or fragments in the original post are covered, and \textit{false} if any parts are missing.
    \item \textbf{Correctness} refers to the accuracy of the explanations provided for each sentence or fragment within a post. It focuses on whether the explanations correctly identify and justify why a fragment is considered hate speech or not. This is evaluated at the \textbf{sentence or fragment level}, with \textit{true} indicating a correct explanation and \textit{false} indicating an incorrect one.
\end{itemize}

The annotators received the guidelines, some examples, and the definition of hate speech. They received a training session and discussed the outcomes, checking and revising the guidelines for any missing aspects. Then, they annotated the full subsample. After annotation, we reported the Inter Annotator Agreement (IAA), showing the percentage of instances where all three annotators fully agreed. For the completeness labelling, we report the percentage of posts where all three annotators provided the same label. For correctness, we report the percentage of fragments within a post where all three annotators have assigned the same label. 

\subsubsection{Annotators}

We recruited three annotators aged 25-30 with diverse backgrounds to evaluate the LLM outputs: one PhD student specializing in hate speech, one psychologist specializing in hate speech and mental health, and another PhD student focused on mental health and Explainable AI. Two annotators are male and one is female. Annotators were presented with labels and explanations in a blind, random setting.

\section{Results}

Next, we present the results regarding both tasks (classification and explanation).

\subsection{Experiment 1: Classification Task}

For evaluating the classification task, we extracted the predicted label for the three LLMs (teacher, small and distilled) and compared it to the golden label. Then, we reported the F1-score, weighted, micro and macro. Moreover, we included four state-of-the-art models with which to compare the hate speech detection task. Table \ref{tab:results} shows the detection results of the three models under evaluation, together with the four baselines, evaluated on ``MiniMetaHate Eval''. 

Among the three LLMs evaluated, the distilled model demonstrates the highest performance, achieving an F1-score of approximately 0.85. This is followed by the Teacher model, which achieves an F1-score of 0.78, representing an approximate 9\% improvement by the distilled model over the teacher model. The smaller model reports an F1-score of less than 0.75, showing the distilled model an enhancement of about 13\% over this baseline. These results highlight the efficacy of our distillation process, which effectively enhances model performance compared to the larger teacher model and the smaller baseline. The improvement of the distilled model over the teacher model is due to the difference in training. The distilled model was fine-tuned for hate speech classification, as well as explanation, using over \num{2000} examples for learning. In contrast, the teacher model performs in a Few-Shot CoT setting, with fewer than \num{20} examples to guide it.

When comparing these results with the selected baselines, it is notable that MetaHateBERT achieves superior performance, with F1-scores exceeding 0.90. However, MetaHateBERT was trained in over one million instances and when training this architecture using the same instances of those used to fine-tune the LLM, the F1 score was about 0.78 ($\sim$7\% of improvement). Moreover, our distilled model surpasses all the other benchmarks. Specifically, it surpasses HateBERT ($\sim$60.4\% of improvement), HateXplain ($\sim$32.9\% of improvement) and the Perspective API ($\sim$11.1\%). 

These findings indicate that the distillation process and the approach used to integrate rationales contribute considerably to improving performance. The strong performance of the distilled model suggests that our method of multi-task learning, where both classification and rationale generation are optimized, provides substantial benefits.

\renewcommand{\arraystretch}{1.3}
\begin{table}[htb]
\centering
\footnotesize
\caption{Hate Speech detection results run on ``MiniMetaHate Eval''.}
\begin{tabular}{ll rrr}
\toprule
& &\textbf{F1} & \textbf{F1\textsubscript{MICRO}} & \textbf{F1\textsubscript{MACRO}} \\ \midrule
\textbf{Llama-3-70B} \textit{(Few-Shot CoT)} & & 0.7796 & 0.7806 & 0.7796   \\ 
\textbf{Llama-3-8B} \textit{(Few-Shot CoT)} & & 0.7467 & 0.7491 & 0.7467   \\
\textbf{Llama-3-8B-Distil-MetaHate} & & \underline{0.8499} & \underline{0.8500} & \underline{0.8499} \\ 
\textbf{MetaHateBERT \cite{Piot_Martin-Rodilla_Parapar_2024}} & & \textbf{0.9037} & \textbf{0.9040} & \textbf{0.9037} \\ 
\textbf{HateBERT \cite{caselli-etal-2021-hatebert}} & & 0.5291 & 0.5917 & 0.5290 \\ 
\textbf{HateXplain \cite{mathewb2020hatexplain}} & & 0.6397 & 0.6662 & 0.6396 \\ 
\textbf{Perspective API \cite{10.1145/3534678.3539147}} & & 0.7650 & 0.7676 & 0.7650 \\ 
\bottomrule
\end{tabular}
\label{tab:results}
\end{table}

\subsection{Experiment 2: Explanation Task}

This task was evaluated using human annotators as described in \cref{sec:annotation}. Table \ref{tab:agreement} shows the percentage of agreement in the manual evaluation of 100 instances. We report the percentage of agreement when all three have annotators assigned the same annotation label. For the correctness attribute, we measured agreement at the phrase level. As shown in Table \ref{tab:agreement}, for the ``Complete'' annotation, all annotators perfectly agreed on 95\% or more for the three models. For the ``Correct'' annotation, the agreement is certainly good, with values around 90\%.

\begin{table}[H]
    \centering
    \footnotesize
    \caption{Inter Annotator Agreement of the manual evaluation of explanations generated.}
    \begin{tabular}{ll rr}
    \toprule
    & & \textbf{Complete} & \textbf{Correct} \\ \midrule
    \textbf{Llama-3-70B} \textit{(Few-Shot CoT)} & & 96.00\% & 90.91\%   \\
    \textbf{Llama-3-8B} \textit{(Few-Shot CoT)} & & 95.00\% & 88.85\%  \\
    \textbf{Llama-3-8B-Distil-MetaHate} & & 95.00\% & 88.55\% \\
    \bottomrule
    \end{tabular}
    \label{tab:agreement}
\end{table}

Next, we present the results based on the majority decision i.e. where two or more annotators agreed on the label. In Table \ref{tab:major_decision}, we can see that both the teacher model and the distilled model achieved a perfect score for completeness, with all instances being complete. The small base model, however, had 97\% complete examples. When evaluating correctness, we report the percentage of posts from the 100-sample subset where all explanations were correct over all fragments of the post. The teacher model performed best, with 92\% of the post explanations being correct, closely followed by the distilled, with 91\%. The small base model had a lower correctness rate, with only 83\% of the posts being correct. This shows that our distilled model is just as effective as the larger language model while being more efficient. Moreover, in Table \ref{tab:error_analysis} we provide an error analysis of both the teacher and distilled models in terms of their classification and explanation performance.

\begin{table}[ht]
    \centering
    \footnotesize
    \caption{Percentage of correct and complete posts on majority decision.}
    \begin{tabular}{ll rr}
        \toprule
        & & \textbf{Complete} & \textbf{Correct} \\ \midrule
        \textbf{Llama-3-70B} \textit{(Few-Shot CoT)} & & \textbf{100\%} & \textbf{92.00\%} \\ 
        \textbf{Llama-3-8B} \textit{(Few-Shot CoT)} & & 97.00\% & 83.00\% \\
        \textbf{Llama-3-8B-Distil-MetaHate} & & \textbf{100\%} &  \underline{91.00\%} \\
        \bottomrule
    \end{tabular}
    \label{tab:major_decision}
\end{table}

\begin{table}[ht]
    \centering
    \footnotesize
    \caption{Comparison of teacher and distil model explanations for hate speech instances.}
    \begin{tabular}{p{3.2cm} p{1.2cm} p{7.5cm}}
        \toprule
        \textbf{Text} & \textbf{Model} & \textbf{Explanation} \\ \midrule
        \multirow{2}{3.2cm}{A word to all the ladies in very short skirts or shorts this summer - 'Thanks'.} & Teacher & \textit{The phrase is hate speech because it objectifies and reduces women to their clothing choices, implying that they are grateful for male attention or approval. It promotes a sexist and patriarchal attitude, where women's bodies are policed and judged.} \\ 
        & Distil & \textit{This phrase is a message of appreciation and gratitude, not promoting hate or discrimination towards any individual or group based on their attire.} \\ \midrule
        \multirow{2}{3.2cm}{I love it when married bitches start posting inspirational memes about having a strong relationship... Single in 3-2-1...} & Teacher & \textit{This phrase is a humorous prediction about the outcome of the relationship, rather than a hateful or discriminatory statement.} \\
        & Distil & \textit{'Married bitches' is a derogatory term used to demean and objectify women, particularly those who are married. This language promotes discrimination and hostility towards women. 'Single in 3-2-1...' is a form of mockery towards single people, implying that they will remain single for a certain period of time.} \\
        \bottomrule
    \end{tabular}
    \label{tab:error_analysis}
\end{table}

In conclusion, our results show that model distillation is effective. The distilled model matches the teacher model in explainability and outperforms it in classifying hate speech. 

In our experiments, we used quantized 4-bit models (\texttt{Llama-3-70B-Instruct} and \texttt{Llama-3-8B -Instruct}) from \url{unsloth}, where the distilled model processes tokens at 0.4143 tokens per second, while the larger model does so at 0.1879 tokens per second ($\sim$54\% slower). The distilled model requires 8.1 GB of GPU memory, compared to 42.5 GB for the larger model. In production, an NVIDIA A100 is needed for the large model, whereas the distilled model runs well on an NVIDIA L4. Environmental impact is lower with the L4: it emits 0.04 kg CO$_2$eq per hour, compared to 0.22 kg CO$_2$eq for the A100. Cost-wise, running an A100 for 30 hours per month on Google Cloud costs \$152.06, while the L4 costs significantly less at \$21.20, for the same duration. The large model requirements limit its practical use, making the distilled model a convenient option for broader applications.

Based on our experimental results, we propose the following hate speech detection pipeline for social media. First, we suggest using MetaHateBERT to classify posts as hate or non-hate speech. Next, the distilled model, \texttt{Llama-3-8B-Dis- til-MetaHate}, will generate explanations for each classification, clarifying the rationale behind the labels. This pipeline aims to improve detection accuracy while enhancing transparency and fairness, contributing to a better online environment.

\section{Conclusion}

In this study, we explored how to detect hate speech in an explainable way on social media and suggested a way to transfer knowledge from large language models to smaller ones. We used a multi-task learning approach, where the models are trained to not only identify hate speech but also provide explanations for their decisions. We evaluated these explanations with humans to check if the generated content agrees with human judgment. Our results show thet the distilled small model, which is faster, greener and more efficient, can both detect and explain hate speech as well as larger models, with solid improvements in classification. Future lines will involve distilling different models to compare with our proposed approach, distilling the model using more examples, and applying other prompting techniques such as Tree-of-Thought (ToT). We also aim to investigate how these methods generalise across different languages and cultural contexts.

\section*{Computational Resources}
Experiments were conducted using a private infrastructure, which has a carbon efficiency of 0.432 kgCO$_2$eq/kWh. A cumulative of 15 hours of computation was performed on hardware of type RTX A6000 (TDP of 300W). Total emissions are estimated to be 1.94 kgCO$_2$eq of which 0 percent were directly offset. Estimations were conducted using the \href{https://mlco2.github.io/impact#compute}{MachineLearning Impact calculator} presented in \cite{lacoste2019quantifying}.

\begin{credits}
\subsubsection{\ackname} The authors thank the funding from the Horizon Europe research and innovation programme under the Marie Skłodowska-Curie Grant Agreement No. 101073351. Views and opinions expressed are however those of the author(s) only and do not necessarily reflect those of the European Union or the European Research Executive Agency (REA). Neither the European Union nor the granting authority can be held responsible for them. The authors also thank the financial support supplied by the Consellería de Cultura, Educación, Formación Profesional e Universidades (accreditation 2019-2022 ED431G/01, ED431B 2022/33) and the European Regional Development Fund, which acknowledges the CITIC Research Center in ICT as a Research Center of the Galician University System and the project PID2022-137061OB-C21 (Ministerio de Ciencia e Innovación supported by the European Regional Development Fund). The authors also thank the funding of project PLEC2021-007662 (MCIN/AEI/10.13039/501100011033, Ministerio de Ciencia e Innovación, Agencia Estatal de Investigación, Plan de Recuperación, Transformación y Resiliencia, Unión Europea-Next Generation EU).

%\subsubsection{\discintname}
%Paloma Piot has received research grants from Horizon Europe research and innovation programme under the Marie Skłodowska-Curie Grant Agreement No. 101073351. Javier Parapar has received financial support by Consellería de Cultura, Educación, Formación Profesional e Universidades (accreditation 2019-2022 ED431G/01, ED431B 2022/33) and the European Regional Development Fund, which acknowledges the CITIC Research Center in ICT as a Research Center of the Galician University System and the project PID2022-137061OB-C21 (Ministerio de Ciencia e Innovación supported by the European Regional Development Fund). Also, Javier Parapar has receive fundins from the project PLEC2021-007662 (MCIN/AEI/10.13039/501100011033, Ministerio de Ciencia e Innovación, Agencia Estatal de Investigación, Plan de Recuperación, Transformación y Resiliencia, Unión Europea-Next Generation EU).
\end{credits}
\bibliographystyle{splncs04}
\bibliography{custom}
\end{document}